\documentclass{article}
\usepackage{spconf,amsmath,graphicx}
\usepackage{rotating}
\usepackage{subfig}
\usepackage{amsmath}

\usepackage[absolute]{textpos}
\newcommand*{\Scale}[2][4]{\scalebox{#1}{$#2$}}%

\usepackage{lipsum}

\let\OLDthebibliography\thebibliography
\renewcommand\thebibliography[1]{
  \OLDthebibliography{#1}
  \setlength{\parskip}{1pt}
  \setlength{\itemsep}{1pt plus 0.3ex}
}

\title{PL-UNeXt: PER-STAGE EDGE DETAIL AND LINE FEATURE GUIDED SEGMENTATION FOR POWER LINE DETECTION}

\name{Yang Cheng\textsuperscript{1} \qquad Zhen Chen\textsuperscript{2} \qquad Daming Liu\textsuperscript{1 $\dagger$} \thanks{$\dagger$ Corresponding Author.}}
  
  \address{\textsuperscript{1}Shanghai University of Electric Power, Shanghai 201101, China\\
      \textsuperscript{2}State Grid Sichuan Electric Power Research Institute, Chengdu 610041, China}

\begin{document}
\begin{textblock}{19}(1.1, 0.6)
\noindent \small Copyright 2023 IEEE. Published in 2023 IEEE International Conference on Image Processing (ICIP), scheduled for 8-11 October 2023 in Kuala Lumpur, Malaysia. Personal use of this material is permitted. However, permission to reprint/republish this material for advertising or promotional purposes or for creating new collective works for resale or redistribution to servers or lists, or to reuse any copyrighted component of this work in other works, must be obtained from the IEEE. Contact: Manager, Copyrights and Permissions / IEEE Service Center / 445 Hoes Lane / P.O. Box 1331 / Piscataway, NJ 08855-1331, USA. Telephone: + Intl. 908-562-3966.
\end{textblock}
%
\maketitle
\begin{abstract}
Power line detection is a critical inspection task for electricity companies and is also useful in avoiding drone obstacles. Accurately separating power lines from the surrounding area in the aerial image is still challenging due to the intricate background and low pixel ratio. In order to properly capture the guidance of the spatial edge detail prior and line features, we offer PL-UNeXt, a power line segmentation model with a booster training strategy. We design edge detail heads computing the loss in edge space to guide the lower-level detail learning and line feature heads generating auxiliary segmentation masks to supervise higher-level line feature learning. Benefited from this design, our model can reach 70.6 F1 score (+1.9\%) on \emph{TTPLA} and 68.41 mIoU (+5.2\%) on \emph{VITL} (without utilizing IR images), while preserving a real-time performance due to few inference parameters.
\end{abstract}
\section{Introduction}
\label{sec:intro}

Detecting power lines is a daily inspection task for electric power companies and remains a tough job. Power lines are also safety threats to UAVs but accurately locating them from aerial images is challenging. Power lines usually measure about a few pixels wide in the image, and the very thin shape can be easily fragmented by other objects.

There have been works utilizing traditional computer vision algorithms and deep learning 
models to detect the power lines. Traditional algorithms suffer a problem of being vulnerable 
to environment changes and most deep learning models are hard to optimize on such a 
specialized task. Combining both advantages may be a way to help. Recently, PLGAN \cite{abdelfattah2022plgan} was proposed to segment power lines from aerial images. PLGAN developed a generated adversarial network structure with a designed Hough transform loss to embed geometry prior into the model. Choi et al.\cite{choi2022attention} proposed an attention-based multimodal feature fusion module to utilize both visual images and infrared images and is capable of segmenting power lines in five light and weather conditions.

 To further exploit the learning ability of neural network models embedded with prior knowledge 
 in power line detection task, we propose to fully utilize the edge and shape prior of the power 
 lines by guiding the model to learn how to separate edges and extract most relevant line features 
 from the learned edges. Our main contributions are summarized as follows:
 \vspace{-1mm}
\begin{itemize}
\itemsep0em
     \item We propose PL-UNeXt, a semantic segmentation model to detect power lines in a real-time 
     speed with booster training strategy to guide the encoder learning edge and line priors.
     \item Edge detail heads with edge space converter modules are proposed to help the lower-level 
     layers of the encoder generate better edge features.
     \item Line feature heads with dynamic line extractors are proposed to adaptively assemble 
     multi-sized line features and lead to more appropriate learning representations from the 
     higher-level layers of the encoder.
\end{itemize}


\begin{figure*}[!t]
    \centering
        \includegraphics[width=1\textwidth]{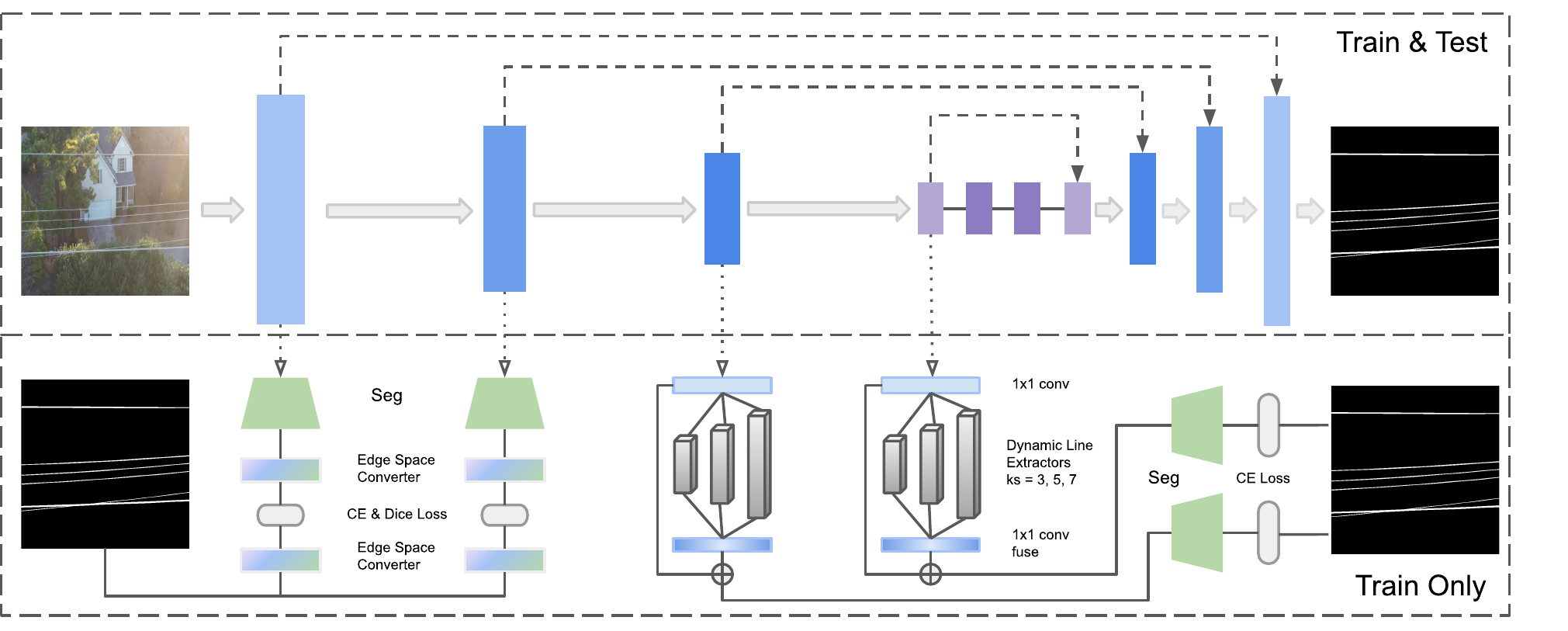}
        \caption{The overview of the PL-UNeXt. The first and second stages of the encoder are guided by the ED heads. The third and fourth stages are guided by the LF heads. The guiding process only exists in the training phase. The Edge Space Converter module and Dynamic Line Extractor module are presented in Fig.\ref{fig:sub-modules}.}
        \vspace{-3mm}
    \label{fig:pl-unext}
\end{figure*}


\section{Related Work}
\label{sec:relate}

\textbf{Booster Training Strategy. } In real-time semantic segmentation, improving the accuracy of
segmentation with same or less inference parameters is challenging. In 
BiSeNetv2 \cite{yu2021bisenet}, a booster training strategy is proposed by inserting
auxiliary segmentation heads such as rocket boosters to enhance the feature representation 
in the training phase and discarded in the inference phase. Additionally, Fan et al. \cite{fan2021rethinking}
proposed STDC with a detail booster branch to guide the lower-level features.

\noindent\textbf{Line Segment Detection. } Line segment detection has similarity to 
power line detection, but a major difference is that we only need power lines. AFM \cite{xue2019learning}
adopted the attraction field map to help calculate the line segments. LCNN \cite{zhou2019end} designed 
a proposal-sampling-verification process to learn relevant line features. HAWP \cite{xue2020holistically} 
designed a wire frame parsing algorithm embedded in an end-to-end learning process. F-Clip \cite{dai2022fully}
reformulated the task of detecting center points and line lengths, and customized line detection blocks with 
bilateral $7\times1$ and $1\times7$ convolutions to exploit line priors.

\section{Proposed Method}
\label{sec:method}

To accurately detect power lines in aerial images while maintaining high speed, 
we propose PL-UNeXt with edge detail guidance and line feature guidance to 
the encoder stages. As shown in Fig.\ref{fig:pl-unext}, PL-UNeXt can be divided into two 
parts, train\&test part containing a UNeXt \cite{valanarasu2022unext} encoder-decoder and the training booster part containing also two parts, the edge detail part with two heads and line feature part with the other two heads. As the power lines occupy a small proportion of the image and have strong shape priors like 
thin, straight, and long, the U-shape network design with a gradually upsample progress layer by 
layer is a suitable choice to recover from higher level feature representations and produce
fine-grained prediction masks.

\subsection{Edge Detail Guidance}
In the work of STDC \cite{fan2021rethinking}, three Laplacian kernels with different
strides were used to extract edge detail from the ground truth mask and then compute the loss 
between the seg logits from the upsampled feature maps and edge details. We explore this further to compute 
loss in the converted edge space to alleviate the asymmetry between the learned convolution feature maps and 
edges from the fixed Laplacian convolutions. The STDC used threshold of $0.1$ to binarize the edges produced by the Laplacian kernels and the gradient becomes impossible to calculate after the threshold operation, so they apply a fixed $1\times1$ convolution kernel to re-weight the three edge maps of different sizes. Here, we abandon the threshold operation and make the process learnable so that the $1\times1$ convolution we used can learn adaptive weights to fuse the edge maps. Also, we replace the three strided Laplacian kernels with three non-strided kernels: a Laplacian kernel, a Sobel X kernel, and a Sobel Y kernel to extract more types of edges, thus can enrich the capability of converting to the proper edge space. 

Suppose $X$ as the seg map of the feature maps from the first two stages, $Y$ as the ground truth, the process can be illustrated as follows:
\setlength{\belowdisplayskip}{1pt} \setlength{\belowdisplayshortskip}{1pt}
\setlength{\abovedisplayskip}{1pt} \setlength{\abovedisplayshortskip}{1pt}
\begin{equation}
\setlength{\jot}{0pt}
\begin{split}
    \Scale[0.9]{\bar{X}=Cat(Laplacian(X),SobelX(X),SobelY(X))} \\
    \Scale[0.9]{\bar{Y}=Cat(Laplacian(Y),SobelX(Y),SobelY(Y))}
\end{split}
\end{equation}
\begin{equation}
\setlength{\jot}{0pt}
\begin{split}
    \Scale[0.9]{\hat{X}=Softmax(Norm(Conv(\hat{X})))} \\
    \Scale[0.9]{\hat{Y}=Softmax(Norm(Conv(\hat{Y})))}
\end{split}
\end{equation}
\begin{equation}
    \Scale[0.9]{Loss=\alpha(CrossEntropy(\hat{X}, \hat{Y}))+\beta(Dice(\hat{X}, \hat{Y}))}
\end{equation}
where $\alpha$ and $\beta$ are super parameters and set to $1.0$ and $0.4$.

\begin{table*}[!ht]
\small
    \centering
    \begin{tabular}{l|c|c c c c c|c|c|c}
    \hline
        Models & Modality & Original & Day & Fog & Night & Snow & Average & Param (M) & Fps \\
        \hline
        UNet \cite{ronneberger2015u}& VL & 60.09 & 57.96 & 56.93 & 55.20 & 57.50 & 57.53 & 31.04 & 32.9 \\
        UNet+EF \cite{choi2019real}& IR+VL & 59.51 & 57.99 & 55.91 & 53.65 & 57.30 & 56.87 & 31.04 & 32.7 \\
        UNet+EF+SE \cite{hu2018squeeze}& IR+VL & 61.13 & 59.75 & 57.56 & 56.33 & 59.55 & 58.84 & 32.44 & 24.9 \\
        UNet+FuseNet \cite{hazirbas2017fusenet}& IR+VL & 60.77 & 59.32 & 57.57 & 55.62 & 58.57 & 58.37 & 12.48 & 22.3 \\
        UNet+MMTM \cite{joze2020mmtm}& IR+VL & 60.95 & 59.61 & 58.08 & 57.39 & 59.20 & 59.04 & 13.53 & 14.5 \\
        UNet+FFM \cite{choi2022attention} & IR+VL & 61.71 & 60.89 & 59.00 & 58.00 & 60.59 & 60.03 & 12.73 & 8.5 \\
        UMFNet \cite{choi2022attention}& IR+VL & 63.31 & 62.47 & 61.59 & 61.14 & 61.57 & 62.01 & 15.70 & 26.0 \\
        UMFNet+FuseNet & IR+VL & 62.85 & 62.05 & 60.79 & 60.47 & 61.53 & 61.54 & 15.70 & 26.1 \\
        UMFNet+MMTM & IR+VL & 63.65 & 62.92 & 62.28 & 61.63 & 62.31 & 62.56 & 16.40 & 18.5 \\
        UMFNet+FFM \cite{choi2022attention}& IR+VL & 64.07 & 63.72 & 62.65 & 62.60 & 63.01 & 63.21 & 15.92 & 11.0 \\
        Ours & VL & \textbf{70.43} & \textbf{70.15} & \textbf{69.57} & \textbf{62.62} & \textbf{69.27} & \textbf{68.41} & 1.47(6.39) & 87.8 \\
        \hline
    \end{tabular}
    \caption{The mIoU results (\%) comparison on VITL dataset~\cite{choi2022attention} with same image size of $256\times256$. The VL and IR in the Modality column means \emph{visible light} and \emph{infrared}. VL indicates that the model used visible image only and IR+VL indicates that the model used both visible image and infrared image. The Ours param number outside the brackets is for the inference stage and the number in the brackets is for the training stage.}
\label{tab:results-vitl-basic}
\vspace{-2mm}
    
\end{table*}

\subsection{Line Feature Guidance}
After processed by the lower-level layers guided by the edge detail prior, higher-level layers need to 
focus more on the real object we need to separate, the power lines. As shown in Fig.\ref{fig:pl-unext}, 
the third and forth stages of the encoder are guided by two line feature heads which extract the line 
features and compute the seg loss with the ground truth labels. In the work of F-Clip \cite{dai2022fully},
a custom line detection block with two parallel convolutions with kernel sizes of $1\times7$ and $7\times1$. We extend this design further to expand the capacity and flexibility of extracting the lines.
The main functional part of the line feature heads is composed of three dynamic line extractors with
different kernel sizes and a $1\times1$ convolution to fuse the extracted features. In dynamic line extractors, we adopt Dynamic Convolution \cite{chen2020dynamic} to increase extractor capability without significant growth of training parameters. In each dynamic line extractor, the feature maps are 
bilaterally processed by two Dynamic Convolutions with kernel size of $1\times N$ and $N\times1$, where $N$ 
is different kernel sizes we set to dynamic line extractors. Then the feature maps are concatenated and processed by $1\times1$ convolution to merge the feature maps in an adaptive way.
Let $X$ be the input feature maps, the process can be illustrated as follows:
\setlength{\belowdisplayskip}{1pt} \setlength{\belowdisplayshortskip}{1pt}
\setlength{\abovedisplayskip}{1pt} \setlength{\abovedisplayshortskip}{1pt}
\begin{equation}
    \Scale[0.9]{Feat=Cat(DyConvX(X),DyConvY(X))}
\end{equation}
\begin{equation}
    \Scale[0.9]{Out=Act(Norm(Conv(Feat)))}
\end{equation}
The outputs of the dynamic line extractors are fused by $1\times1$ convolution and concatenated 
with the original feature maps. Then the final outputs are used to generate train-only predictions 
to calculate the auxiliary segmentation loss.

\subsection{Optimization with Auxiliary Loss}
Our proposed booster strategy exists only in the training phase and all boosters are discarded in the testing phase. During training, the losses of the booster heads will conduct the gradient 
to the backbone of the model which exists all time. The overall loss during training is as follows:
\setlength{\belowdisplayskip}{1pt} \setlength{\belowdisplayshortskip}{1pt}
\setlength{\abovedisplayskip}{1pt} \setlength{\abovedisplayshortskip}{1pt}
\begin{equation}
    \Scale[0.9]{L=\theta L_{decode}+\iota L_{ED1} + \kappa L_{ED2} + \lambda L_{LF1} + \mu L_{LF2}}
\end{equation}
where $\theta$, $\iota$, $\kappa$, $\lambda$ and $\mu$ are all set to $1.0$.

\begin{figure}[t]
\centering

\subfloat[Edge Space Converter]{%
  \includegraphics[width=3.3cm]{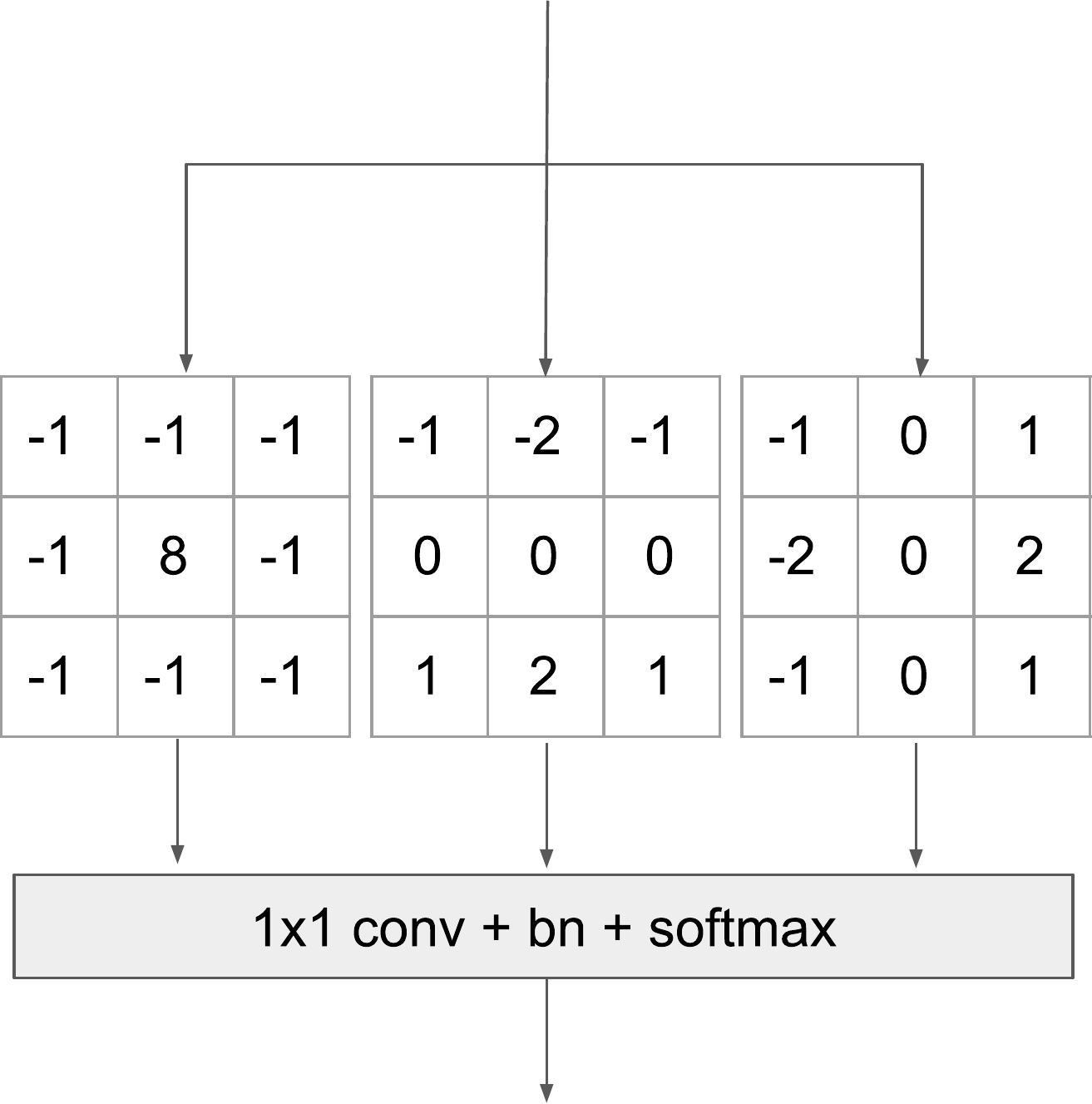}%
  \label{fig:esc}%
}\qquad
\subfloat[Dynamic Line Extractor]{%
  \includegraphics[width=3.3cm]{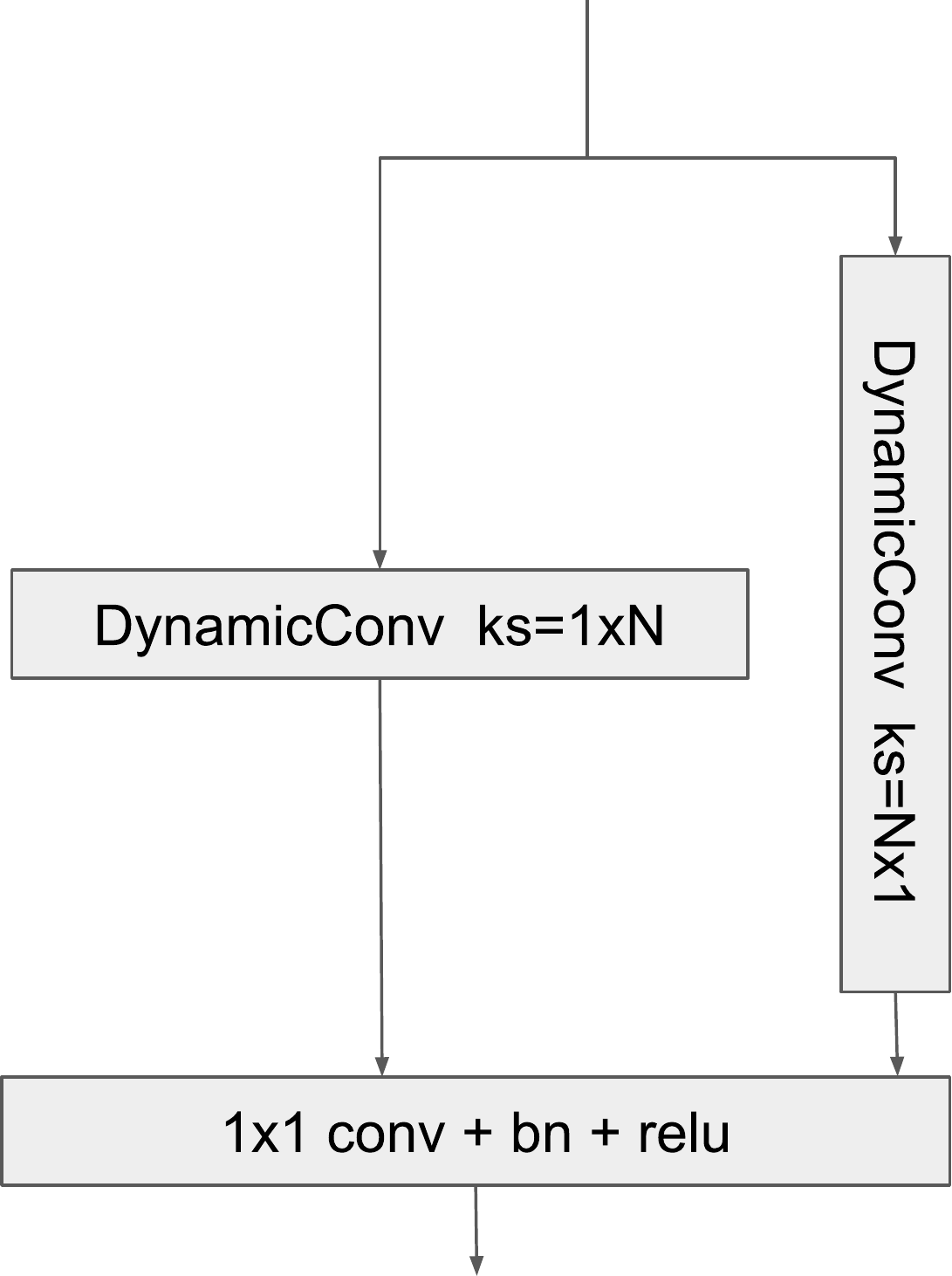}%
  \label{fig:dle}%
}

\caption{Structure of the sub-modules. }
\label{fig:sub-modules}
\end{figure}

\section{Experiments}
\label{sec:experiments}

\subsection{Implementation Details}
\label{ssec:impldetail}

\textbf{Dataset and Metric.} We evaluate our method on TTPLA and VITL following 
the settings of PLGAN\cite{abdelfattah2022plgan} and UMFNet\cite{choi2022attention} 
and also compare the results with them as baselines. We adopt F1 score, Precision, 
Recall and IoU for the evaluation on TTPLA, and mIoU for the test on VITL, following
the way they presented.

\noindent\textbf{Settings and Platform.} To train our model, we choose AdamW as the optimizer 
with an initial learning rate of 0.0005 and weight decay of 0.05. We use CosineAnnealing
policy with a minnimum learning rate of 5e-6 to find more optimal minima. Our method is 
implemented on Pytorch and the experiments are run on NVIDIA GeForce RTX 3050.

\begin{figure*}[!ht]
        \centering

        \setlength{\tabcolsep}{2pt}
        \begin{tabular}{cccccc}

        \includegraphics[width=2.4cm]{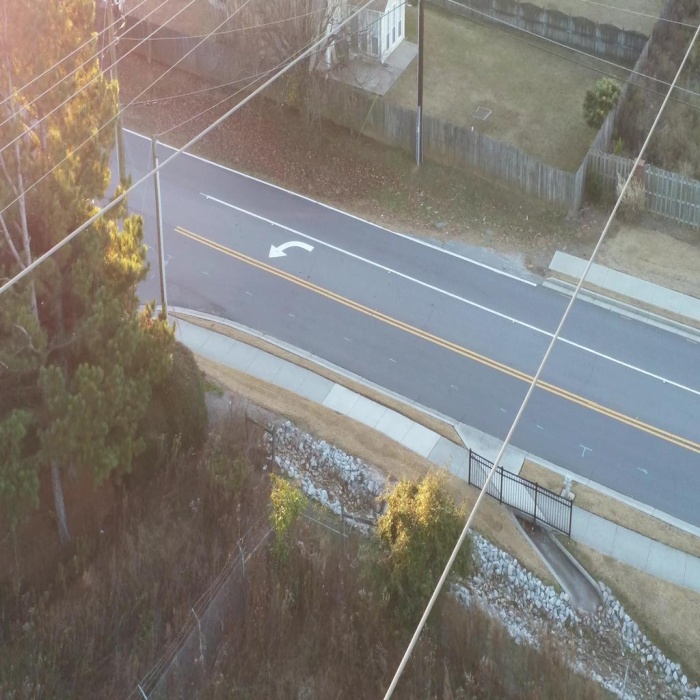}&
        \includegraphics[width=2.4cm]{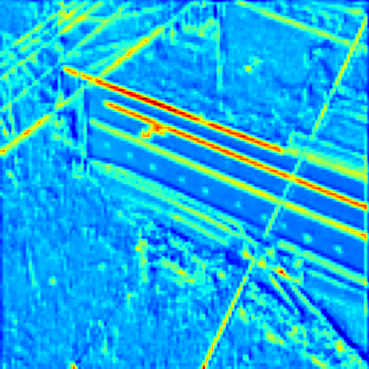}&
        \includegraphics[width=2.4cm]{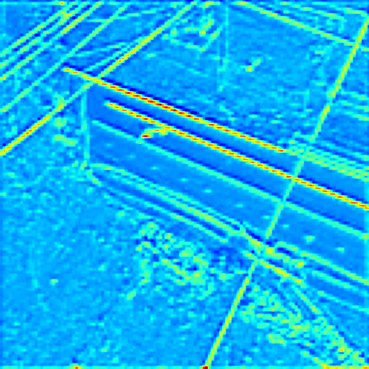}&
        \includegraphics[width=2.4cm]{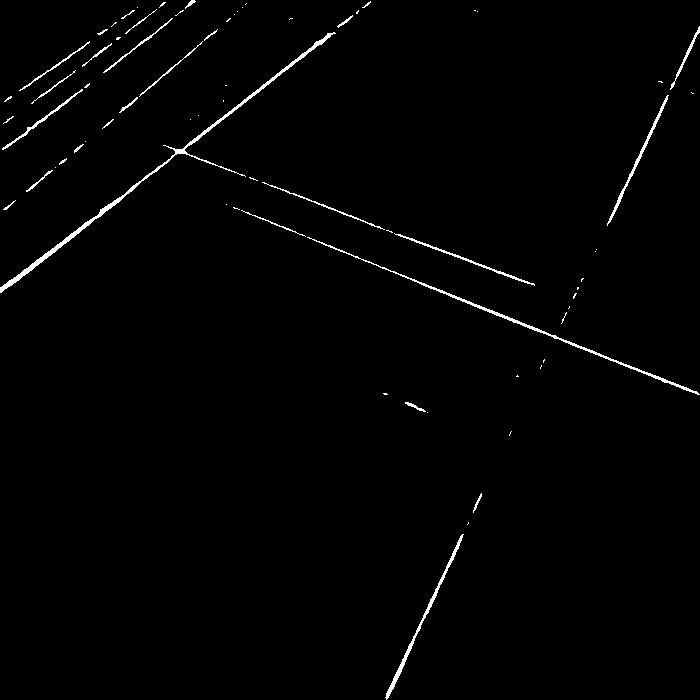}&
        \includegraphics[width=2.4cm]{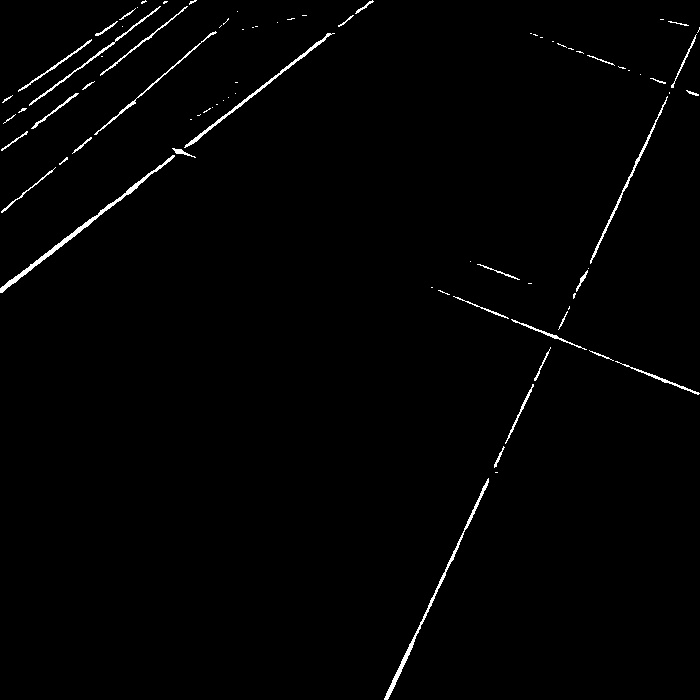}&
        \includegraphics[width=2.4cm]{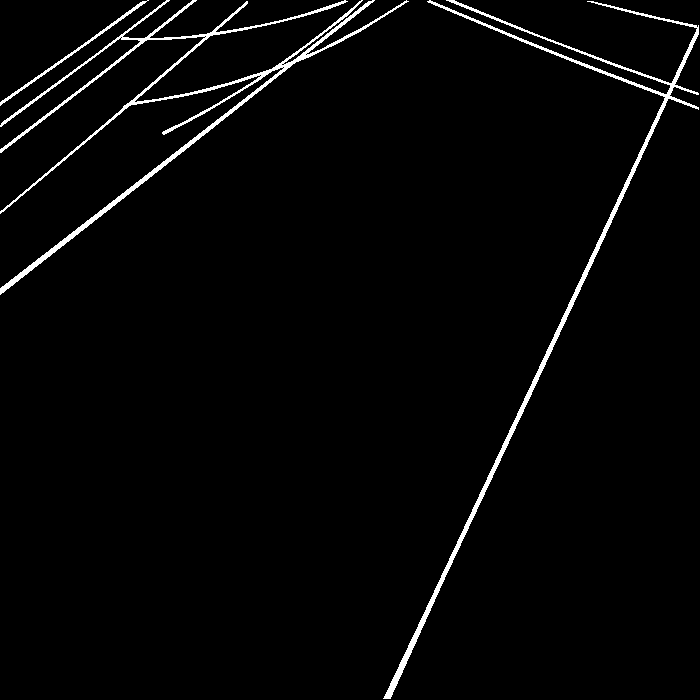} \\
        
        \includegraphics[width=2.4cm]{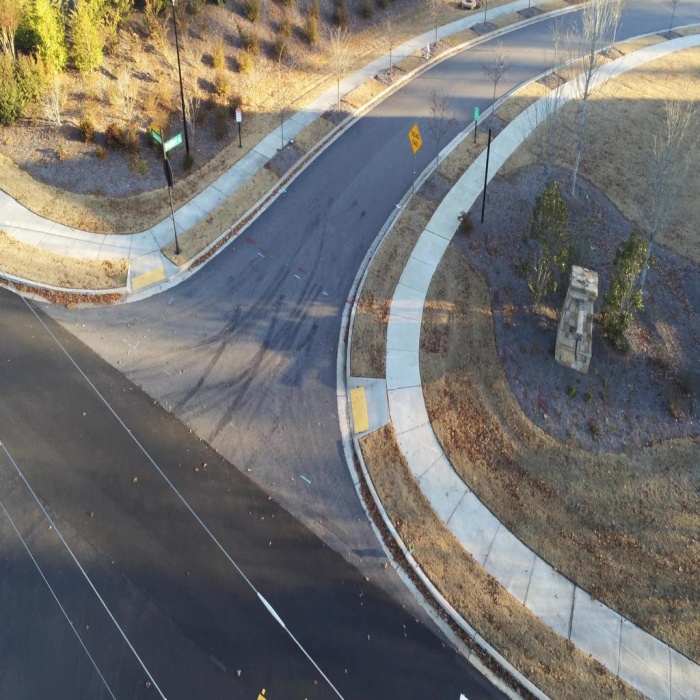}&
        \includegraphics[width=2.4cm]{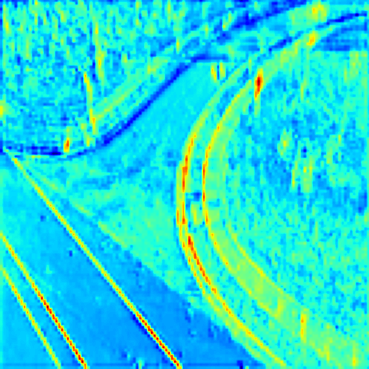}&
        \includegraphics[width=2.4cm]{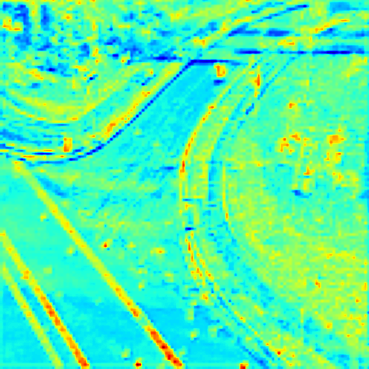}&
        \includegraphics[width=2.4cm]{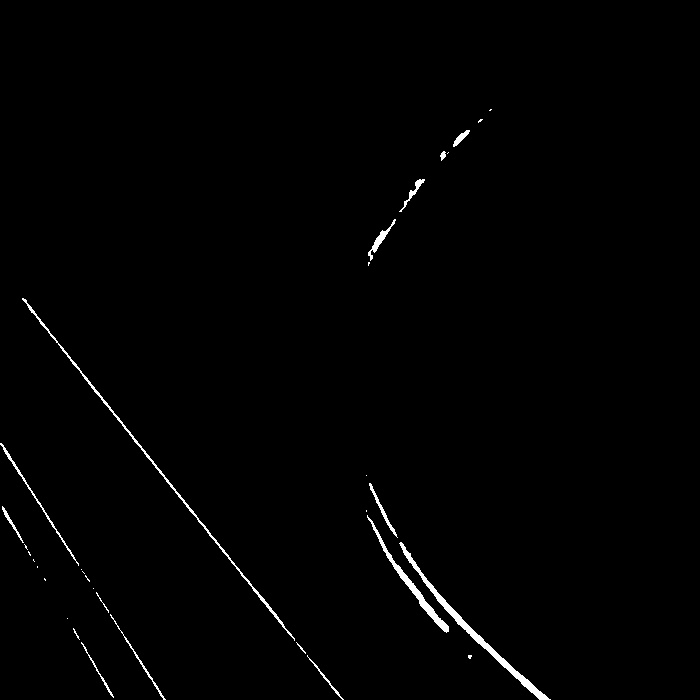}&
        \includegraphics[width=2.4cm]{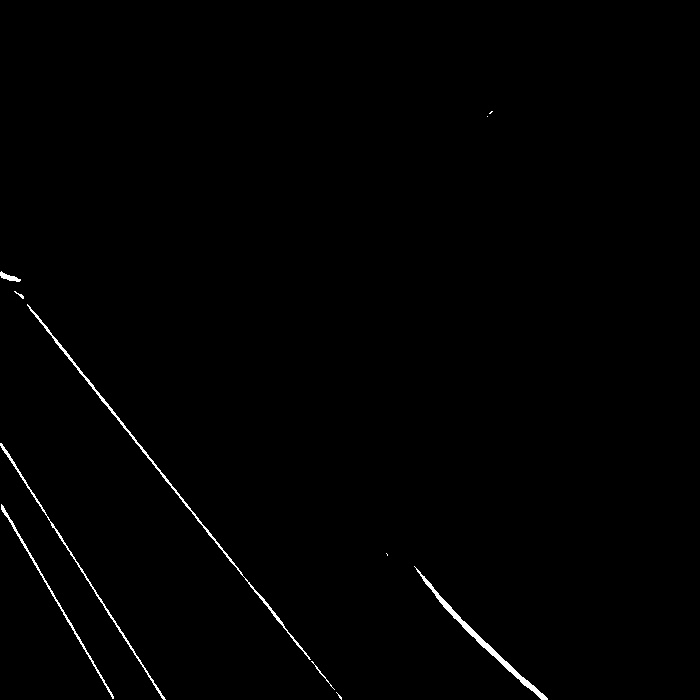}&
        \includegraphics[width=2.4cm]{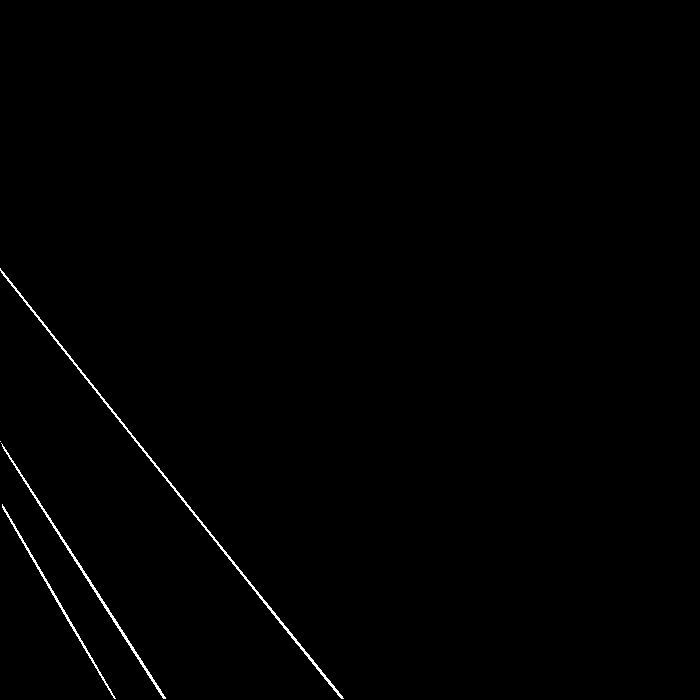} \\
        
        \includegraphics[width=2.4cm]{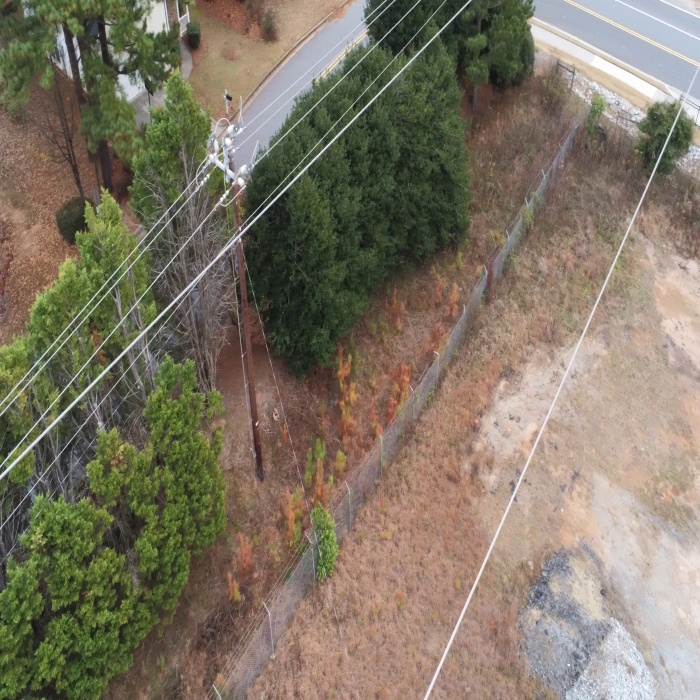}&
        \includegraphics[width=2.4cm]{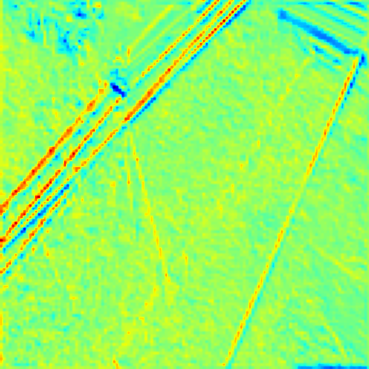}&
        \includegraphics[width=2.4cm]{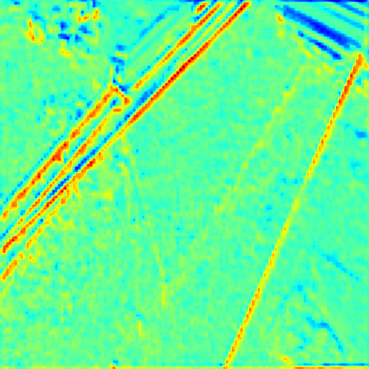}&
        \includegraphics[width=2.4cm]{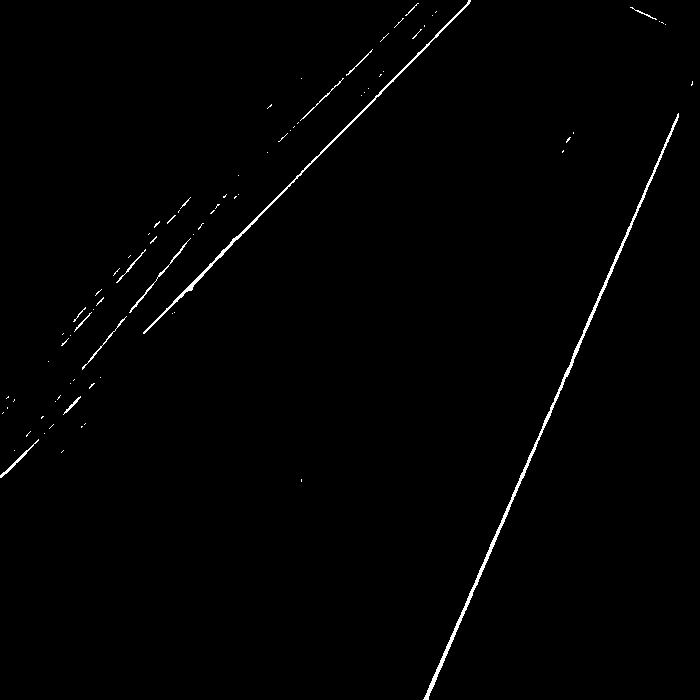}&
        \includegraphics[width=2.4cm]{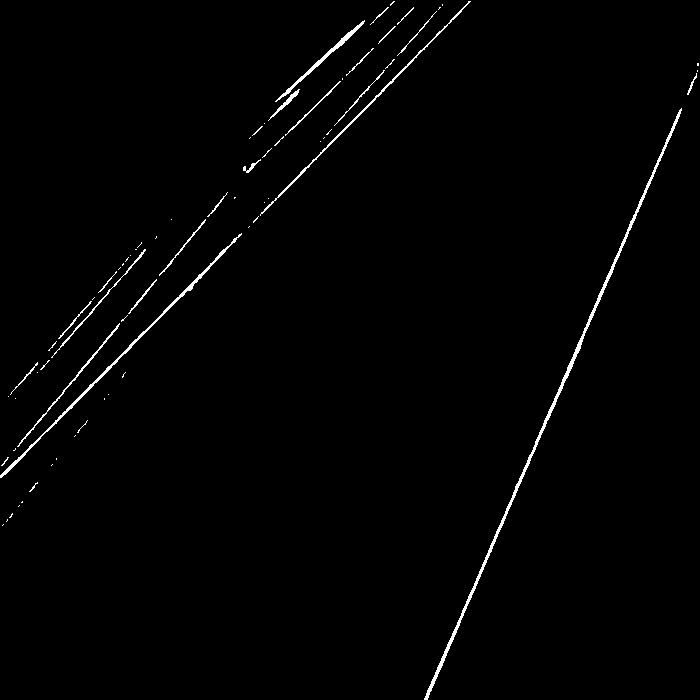}&
        \includegraphics[width=2.4cm]{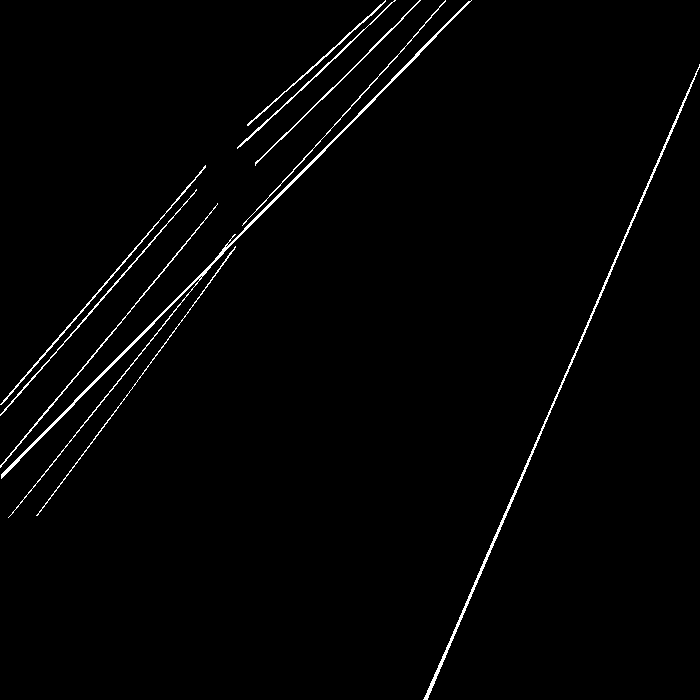} \\

        \footnotesize Input &
        \footnotesize Stage3 &
        \footnotesize Stage3$_G$ &
        \footnotesize Prediction &
        \footnotesize Prediction$_G$ &
        \footnotesize GroundTruth \\
        \end{tabular}
        \caption{Visual comparison of the feature maps and predictions on TTPLA. The subscript $G$ denotes results with guidance of our designed edge detail and line feature heads. }
        \vspace{-0.5mm}
        \label{fig:visualizations}
\end{figure*}

\subsection{Experimental Results}
\label{ssec:exprresult}

\textbf{TTPLA. } As shown in Tab.\ref{tab:results-ttpla-basic}, our PL-UNeXt outperforms PLGAN 
with minor superiority. The F1 score and IoU improves by 1.9\% and 1.3\%. The Precision rate of 
PLGAN is still the highest but our Recall rate outperforms by 5.7\%, contributing to a more balanced 
detecting performance.

\noindent\textbf{VITL. } Considering not all UAVs are equipped with infrared cameras, our model is 
proposed to a more generalized scene with regular cameras so we use visible images only. As shown in 
Tab.\ref{tab:results-vitl-basic}, our model outperforms by 5.2\% mIoU in the average value of the 
five light and weather conditions preserving the highest fps. 

\noindent\textbf{Visualization. } We visualize the feature maps produced by stage 3 of the encoder and 
final prediction masks in Fig.\ref{fig:visualizations}. The visualizations compare the results produced without and with edge detail and line feature guidance. The feature maps show that the weights are more focused and 
neat on the power lines, thus contributing to predictions closer to ground truths and less false positives.

\begin{table}[!ht]
\small
    \centering
    \begin{tabular}{l|c|c|c|c|c}
    \hline
        Models & $F_1$ & IoU & Pre. & Rec. & P (M) \\
        \hline
        FPN~\cite{lin2017feature} & 0.569 & 0.423 & 0.769 & 0.513  & 23.2   \\
        UNET~\cite{ronneberger2015u}\ & 0.662 & 0.515 & 0.846 & 0.583  & 24.4   \\
        UNET++ \cite{zhou2018unet++}\ & 0.668 & 0.522 & 0.843 & 0.591  & 26.1  \\
        Pix2pix~\cite{isola2017image} & 0.663 & 0.509 & 0.822 &0.577 &10.6   \\
        GcGAN~\cite{fu2019geometry} & 0.655 & 0.501 & 0.837 & 0.556 &13.4   \\
        AFM~\cite{xue2019learning} & 0.457 & 0.307 & 0.495 & 0.432 &44.0  \\
        LCNN~\cite{zhou2019end}& 0.498 & 0.315  & 0.541  & 0.464    &10.9    \\
        HAWP~\cite {xue2020holistically} &0.485 & 0.315   & 0.581  & 0.421  & 11.6   \\
        PLGAN~\cite{abdelfattah2022plgan} & 0.687 & 0.533 &\textbf{0.863}  & 0.577  & 14.9  \\ 
        Ours & \textbf{0.706} & \textbf{0.546} & 0.821 & \textbf{0.634} & 1.5 \\
        \hline
    \end{tabular}
    \caption{PL segmentation performance of the proposed and the comparison methods on TTPLA dataset~\cite{abdelfattah2020ttpla}. Following previous works, models are evaluated on the images and labels resized to $512\times512$. The number of params is for inference stage.}
\label{tab:results-ttpla-basic}
    
\end{table}
\vspace{-2mm}

\subsection{Ablation Study}
\label{ssec:ablation}

To prove the effectiveness of our designed ED and LF, we separately added ED and LF heads on the first two 
stages and the last two stages of the encoder. The results are presented in Tab.\ref{tab:results-ttpla-ablation} and can be seen that the ED and LF achieved 1.1\% F1 and 1.4\% F1 score improvements, also 1.2\% and 1.6\% 
improvements on IoU. Altogether the two ED and LF contributed the final proposed model and reached higher accuracy. The visualization comparison in Fig.\ref{fig:visualizations} clearly shows that our ED and LF can help guide the encoder by focusing on more relative line shapes and producing more precise predictions.

\begin{table}[!ht]
\small
    \centering
    \begin{tabular}{l|c|c}
    \hline
        Methods & $F_1$ & IoU  \\
        \hline
        Base & 0.683 & 0.519 \\
        Base + ED & 0.694 & 0.531 \\
        Base + LF & 0.697 & 0.535 \\
        Base + ED + LF & 0.706 & 0.546 \\
        \hline
    \end{tabular}
    \caption{Ablation study of the methods on TTPLA dataset~\cite{abdelfattah2020ttpla}. Here the Base is the UNeXt network, ED is the proposed edge detail head and LF is the proposed line feature head.}
\label{tab:results-ttpla-ablation}
    
\end{table}
\vspace{-2mm}

\section{Conclusion}
\label{sec:conclusion}
In this work, we propose PL-UNeXt with a booster training strategy to maximize edge and shape prior knowledge in power line segmentation. We first propose 
two edge detail heads to guide the lower-level layers of the encoder to learn 
better edge proposals, then we propose two line feature heads to guide the higher-level 
layers of the encoder by adaptively extracting and verifying the relevant line feature representations.
Extensive experiments show that PL-UNeXt outperforms recently proposed methods. Visualizations 
show that our design is reasonable to fit the need of our envisage.

\vfill\pagebreak

\bibliographystyle{IEEEbib}
\bibliography{strings,refs}

\end{document}